\documentclass{amia}
\usepackage{lipsum} 
\usepackage{xcolor}
\usepackage{graphicx}
\usepackage{booktabs}
\usepackage{multirow}
\usepackage{enumitem}   
\usepackage{wrapfig}
\usepackage{amsmath}
\begin{document}

\title{A Comparative Study of Feature Selection Methods for EHR Diagnosis Codes in Opioid Use Disorder Prediction}

\author{
Zihan Ding, M.S.$^{1}$, 
Yinan Liu$^{1}$, 
Tengfei Ma, Ph.D.$^{1}$, 
Rachel Wong, M.D., M.P.H., M.B.A., M.S.$^{1,2}$, 
Xia Zhao, M.S.$^{2}$,
Richard N. Rosenthal, M.D.$^{1,2\dagger}$, 
and Fusheng Wang, Ph.D.$^{1\dagger}$\footnotemark[1]
\footnotetext[1]{Both are corresponding authors.}
}

\institutes{ \vspace{2mm}
$^{1}$Stony Brook University,  
$^{2}$Stony Brook Medicine
}

\maketitle

\section*{Abstract}
\textit{Feature selection is a critical step in electronic health record (EHR)-based predictive modeling, where input variables are often high-dimensional, sparse, noisy, and redundant. Large feature sets not only increase computational burden and overfitting risk, but also make model interpretation difficult, leading to limited usefulness in clinical settings. In this study, we focus on diagnosis-related features and compare five feature selection paradigms for opioid use disorder (OUD) prediction: recurrence enrichment, NTK-motivated early gradient sensitivity, LightGBM-SHAP, Elastic Net, and large language model (LLM)-guided semantic selection. We use a unified preprocessing and evaluation framework and assess each method by downstream predictive performance, resampling stability, and representation of infrequent diagnosis codes. Our results demonstrate that performance improves with larger feature budgets with diminishing returns beyond a moderate size.  NTK sensitivity provides the best overall balance of accuracy and stability, and LLM-guided selection contributes complementary clinically meaningful signals despite lower standalone performance.}


\section*{Introduction}

The United States continues to face a severe opioid crisis, with an estimated ten million people aged 12 and over using opioids and an average of 130 overdose deaths per day in 2019 \cite{samhsa2020,cdc_training_2025}.
To address this public health challenge, electronic health record (EHR)-based predictive modeling has become a core methodology  to identify patients at elevated risk of opioid misuse, dependence, and overdose.\cite{rasmy2020representation,dong2020machine,dong2021predicting,dong2023integrated,ding2025hibert}
Within structured EHR data, diagnosis codes are especially attractive because they are routinely collected, clinically interpretable, and standardized enough to support reuse across studies and institutions.
As a result, diagnosis-code features are widely used to summarize patient morbidity burden in downstream prediction tasks.
However, raw diagnosis-code representations are difficult to use directly.
At the patient level, they are extremely sparse and high-dimensional; at the system level, they are influenced by coding conventions, terminology mappings, and site- or time-specific documentation practices.\cite{beam2018big,rajkomar2018scalable}
In addition, diagnosis-code frequencies are strongly imbalanced, with a small set of common codes accounting for much of the observed signal while many codes are rarely observed. 
These characteristics make the full diagnosis-code space noisy, redundant,  difficult to interpret, and potentially unstable for predictive modeling,  even before any classifier is trained.\cite{choi2016doctor}

In this setting, feature selection is not merely a convenience for computational efficiency; it is a central methodological problem. \cite{guyon2003introduction}
Classical frameworks distinguish filter, wrapper, and embedded methods; however, their behavior in high-dimensional diagnosis-code spaces remains poorly understood.\cite{tibshirani1996regression}
Diagnosis-code representations typically contain thousands of candidate variables that are sparsely observed and correlated through coding hierarchies and comorbidity patterns\cite{miotto2016deep,rajkomar2018scalable}, while clinically meaningful signals may reside in relatively rare, phenotype-specific codes rather than globally prevalent ones.
Consequently, simple frequency filtering or purely univariate rankings may discard rare but important diagnoses, whereas embedded models may preferentially select dominant or unstable surrogate features among correlated codes.
Although stability-oriented ideas have been proposed in the statistical literature\cite{nogueira2018stability}, systematic evaluations of feature-selection strategies for diagnosis-code–based EHR modeling remain limited, particularly in settings that simultaneously consider predictive performance, resampling stability, and representation of rare clinical codes.

Prior work in EHR machine learning has largely emphasized representation learning and temporal prediction rather than explicit subset selection over diagnosis codes.
The Deep Patient framework demonstrated that unsupervised representations learned from large-scale EHR data can improve disease prediction across diverse outcomes.\cite{miotto2016deep}
Doctor AI and RETAIN then showed how longitudinal diagnosis sequences can be modeled for future event prediction and interpretable clinical forecasting, respectively\cite{choi2016doctor}.
More directly relevant to feature discovery, Hong et al. introduced KESER in 2021, which uses code embeddings and sparse regression to identify related codified features without requiring patient-level data sharing.\cite{hong2021clinical}
Ghasemi and Lee compared several unsupervised feature-selection methods on ICD and ATC code sets for coronary heart disease, highlighting both performance and interpretability trade-offs in 2024.\cite{ghasemi2024unsupervised}
More recently, large language models (LLMs) have been explored as knowledge-guided tools for structured EHR phenotyping, where they can propose clinically relevant criteria for phenotype construction. \cite{yan2024large}
Outside healthcare-specific benchmarking, emerging machine learning work further suggests that LLMs can rank predictive variables competitively even without direct access to downstream training data.\cite{jeong2024llm} 
Taken together, this literature motivates a broader comparison of statistical, machine-learning, and LLM-assisted approaches for selecting compact diagnosis-code feature sets from sparse EHR data.

To address this gap, we conduct a systematic comparison of multiple feature-selection strategies for diagnosis-code features derived from EHR data. 
The evaluated approaches span three methodological categories: statistical ranking methods, machine-learning-based embedded selection methods, and LLM-assisted feature identification. 
All methods are implemented within a unified preprocessing and modeling pipeline, ensuring that differences in downstream performance arise from the feature-selection strategy rather than inconsistencies in data processing or model configuration.
Beyond conventional predictive metrics, we further analyze qualitative properties of the resulting feature sets, including stability under resampling and the extent to which different methods retain infrequent (long-tail) diagnosis codes.
By jointly examining discrimination, feature compactness, selection stability, and representation of rare codes, this study provides a systematic empirical assessment of diagnosis-code feature selection and highlights practical trade-offs relevant to building robust EHR-based prediction models.

\section*{Methods}

\begin{figure}[t]
\centering
\includegraphics[width=0.9\linewidth]{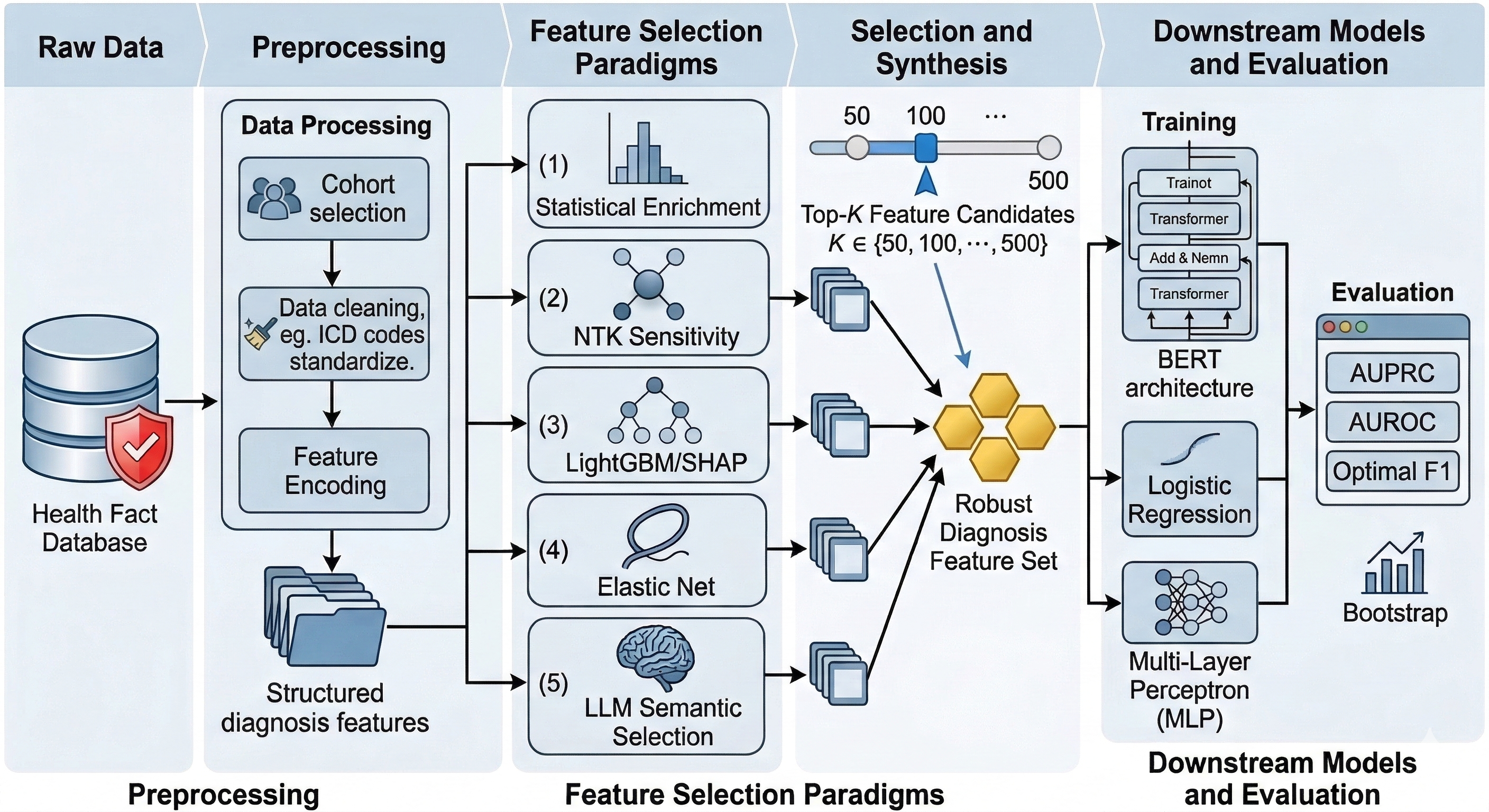}
\caption{Feature Selection and Evaluation Pipeline}
\label{fig:bootstrap}
\end{figure}


We compared five feature selection methods that reflect different inductive assumptions about what makes a diagnosis feature predictive in high-dimensional, sparse EHR data: 
(1) statistical enrichment, 
(2) early gradient sensitivity, 
(3) tree-based model importance, 
(4) regularized linear modeling, and 
(5) semantic-guided selection.
These methods were selected to represent distinct and commonly used paradigms for feature selection.
To enable controlled comparison, each method produced a ranked list of candidate diagnosis features. For downstream modeling, we constructed top-$K$ feature subsets from each ranking under identical data splits and evaluation protocols.

\begin{itemize}
    \item \textbf{Recurrence-Aware Case Enrichment (Recurrence Enrichment).}
    This filter-based approach prioritizes diagnosis features that are 
    over-represented among OUD-positive patients, while accounting for 
    recurrence across encounters. Let $\mathcal{P}^+$ and $\mathcal{P}^-$ 
    denote the sets of OUD-positive and OUD-negative patients, with 
    $N^+ = |\mathcal{P}^+|$ and $N^- = |\mathcal{P}^-|$. For each patient 
    $p$ and diagnosis $d$, we define
    \begin{equation}
        f_d(p) \;=\; \text{number of distinct encounters in which diagnosis } 
        d \text{ appears for patient } p,
    \end{equation}
    which captures diagnosis recurrence across encounters rather than 
    treating occurrence as binary. The population-normalized mean recurrence 
    for each patient group is then computed as
    \begin{equation}
        \bar{f}_d^{+} = \frac{1}{N^{+}}\sum_{p \,\in\, \mathcal{P}^{+}} f_d(p),
        \qquad
        \bar{f}_d^{-} = \frac{1}{N^{-}}\sum_{p \,\in\, \mathcal{P}^{-}} f_d(p).
    \end{equation}
    Diagnoses are ranked by the enrichment score
    \begin{equation}
        \Delta_d \;=\; \bar{f}_d^{+} - \bar{f}_d^{-},
    \end{equation}
    with a higher $\Delta_d$ indicating greater recurrence elevation in 
    OUD-positive patients. Only diagnoses meeting a minimum patient-support 
    threshold ($n_d \geq 50$) were retained. Features were ranked by 
    $\Delta_d$ in descending order, yielding a model-agnostic ranking 
    driven solely by observed outcome associations.

    \item \textbf{NTK-Motivated Early Gradient Sensitivity (NTK Sensitivity).}
    This method ranks diagnosis features by their gradient magnitude at model 
    initialization, without iterative training. For each patient $p$ and 
    diagnosis $d$, the recurrence-weighted feature is defined as
    \begin{equation}
        x'_d(p) \;=\; \log\bigl(1 + f_d(p)\bigr),
    \end{equation}
    where $f_d(p)$ is the encounter-level recurrence count defined in 
    Equation~(1). Under a logistic regression initialized with bias 
    $b_0 = \log\!\left({r}/{(1-r)}\right)$, where $r$ is the 
    population-level OUD prevalence, the predicted probability at 
    initialization is $\hat{p}_0 = \sigma(b_0)$. For each encounter $e$ 
    of patient $p$ with label $y \in \{0,1\}$, the gradient contribution 
    of feature $d$ is
    \begin{equation}
        g_d(e) \;=\; (\hat{p}_0 - y)\cdot x'_d(p).
    \end{equation}
    Diagnoses are ranked by their aggregated absolute gradient across all 
    encounters,
    \begin{equation}
        S_d \;=\; \sum_{e} \bigl|g_d(e)\bigr|,
    \end{equation}
    with larger $S_d$ indicating greater sensitivity at initialization and, 
    by NTK theory, greater expected influence during subsequent training.

    \item \textbf{Gradient Boosting with SHAP Attribution (LightGBM--SHAP).}
    We trained a gradient boosting decision tree model (LightGBM) on 
    encounter-level binary diagnosis features, with patient-level grouping 
    enforced during training and validation splitting to prevent information 
    leakage. Feature importance was quantified via SHAP (SHapley Additive exPlanations) values $\phi_d(e)$, which measure each feature's marginal contribution to the model output for a given encounter $e$. Diagnoses 
    were ranked by their mean absolute SHAP value over the validation set,\cite{lundberg2017unified}
    \begin{equation}
        \bar{\phi}_d \;=\; \frac{1}{|\mathcal{E}_\text{val}|}
        \sum_{e\,\in\,\mathcal{E}_\text{val}} \bigl|\phi_d(e)\bigr|,
    \end{equation}
    with larger $\bar{\phi}_d$ indicating greater average contribution to 
    OUD predictions. This paradigm captures nonlinear effects and 
    conditional dependencies among sparse diagnosis features that linear 
    models may miss.
    
    \item \textbf{Elastic Net Regularized Logistic Regression (Elastic Net).}
    We fitted a logistic regression model with Elastic Net regularization,
    \begin{equation}
        \min_{\boldsymbol{\beta}} \;\mathcal{L}(\boldsymbol{\beta}) 
        + \lambda\!\left[\frac{1-\alpha}{2}\|\boldsymbol{\beta}\|_2^2 
        + \alpha\|\boldsymbol{\beta}\|_1\right],
    \end{equation}
    where $\mathcal{L}$ is the binary cross-entropy loss, $\alpha \in [0,1]$ 
    controls the L1/L2 trade-off, and $\lambda > 0$ governs overall penalty 
    strength. Both hyperparameters were selected by grid search based on 
    validation AUROC. Diagnosis features were ranked by $|\hat{\beta}_d|$, 
    the absolute value of the fitted coefficient, with the L1 component 
    inducing exact sparsity and the L2 component stabilizing estimates under 
    feature correlation.
    
    \item \textbf{Large Language Model--Guided Semantic Prior (LLM Semantic).}
    We employed a two-stage LLM-based strategy using Claude Sonnet 4.5 (Anthropic; temperature = 0). In the first stage, each candidate's ICD-10 three-digit code was evaluated independently via a structured prompt \ref{fig:llm_prompts} that provided the code description and aggregated cohort-level OUD prevalence statistics; no individual patient records were shared with the model at any stage, ensuring compliance with patient privacy requirements. The model was instructed to assess clinical plausibility as an OUD precursor and to explicitly exclude codes representing consequences or treatments of established OUD to prevent label leakage. In the second stage, all endorsed codes were reviewed collectively, and the model selected the final top 100 features based on clinical relevance, specificity, and non-redundancy.
\end{itemize}

\begin{figure}[h]
\centering
\fbox{%
  \begin{minipage}{0.92\textwidth}
    \small
    \textbf{Stage 1 — Per-code evaluation prompt}\\[4pt]
    Evaluate if the following ICD-10 diagnosis code should be included as a 
    feature for predicting future Opioid Use Disorder (OUD).\\[6pt]
    \texttt{ICD-10: \{code\} — \{description\}}\\
    \texttt{OUD patients: \{n$_+$\} \;|\; Non-OUD: \{n$_-$\} \;|\; OUD rate: \{\%\}}\\[6pt]
    EXCLUDE if the diagnosis is a consequence or treatment of established OUD 
    (i.e., label leakage). INCLUDE if it is clinically linked to pain management, 
    opioid exposure, or substance use vulnerability.\\[6pt]
    Reply in JSON only:\;
    \texttt{\{"include": true/false,\; "confidence": 0.0--1.0,\; "rationale": "<50 words>"\}}

    \vspace{10pt}
    \hrule
    \vspace{8pt}

    \textbf{Stage 2 — Top-100 shortlisting prompt}\\[4pt]
    From the \{N\} ICD-10 codes endorsed in Stage 1, select the \textbf{TOP 100} 
    most clinically meaningful categories for OUD risk prediction. Apply the 
    following criteria in priority order:\\[4pt]
    \begin{enumerate}[noitemsep, leftmargin=1.5em]
      \item Strong relevance to opioid exposure, pain management, substance use, 
            or mental health conditions commonly associated with OUD.
      \item Preference for broader clinical concepts over narrow or administrative codes.
      \item Non-redundancy: if multiple codes reflect the same concept, 
            retain the most representative one.
      \item Do \textit{not} base decisions purely on frequency counts; 
            Prioritize clinical judgment.
    \end{enumerate}
    \vspace{4pt}
    Output valid JSON only: a list of 100 objects, each with fields 
    \texttt{ICD10}, \texttt{DIAGNOSIS\_DESCRIPTION}, and \texttt{rationale}.
  \end{minipage}%
}
\caption{Simplified prompts used in the two-stage LLM-based feature selection procedure. 
Placeholders in curly braces were populated with code-specific values at runtime. 
No individual patient records were included; only aggregated cohort-level statistics 
were provided.}
\label{fig:llm_prompts}
\end{figure}

For each method, we constructed feature subsets under multiple budgets,
$K \in \{50, 100, 150, 200, 300, 500\}$, enabling analysis of the performance--complexity trade-off across degrees of dimensionality reduction.

\subsection*{\textit{Data Source}}
We extracted de-identified EHR data from the Cerner Health Facts database, a large multi-institutional repository containing routinely collected clinical data, including patient demographics, encounters, diagnoses, medications, procedures, laboratory results, and administrative/billing information.\cite{cerner2018,deshazo2015comparison}

To define the study cohort, we included patients with at least one recorded prescription of an opioid medication in the medication records. The cohort construction was anchored on opioid exposure, such that each patient contributed longitudinal diagnosis history associated with their observed care trajectory in Health Facts. For feature construction, diagnosis records were restricted to encounters occurring within the 12 months preceding the index opioid prescription, with at most five encounters retained per patient. The final cohort included 11,791,858 patients, of whom 137,214 (1.16\%) were identified as having OUD. The mean age was 48.4 ± 21.9 years, and 58.2\% of patients were female.

For feature construction, we used diagnosis codes stored in EHR. Because the study period spans transitions in coding practices and may contain both ICD-9-CM and ICD-10-CM codes, we harmonized diagnoses by mapping all ICD-9-CM codes to ICD-10-CM, thereby ensuring a unified diagnosis-code space for downstream modeling and feature selection.\cite{drugbank5.0}
To mitigate excessive granularity and reduce fragmentation induced by the ICD codes hierarchy, we further truncated ICD-10-CM codes to the first three characters (category-level representation). This produces a more compact and comparable feature vocabulary (e.g., collapsing subcodes within the same parent category), while retaining clinically meaningful groupings.
Opioid use disorder (OUD) was operationalized using the ICD-10-CM category F11.xx\cite{shickel2017deep}. Patients were considered to have evidence of OUD if an F11-prefixed diagnosis appeared in their diagnosis records.\cite{moore2017}

\subsection*{\textit{Downstream Model}}
To evaluate the clinical utility of each feature selection method, we employed a BERT-based transformer encoder model trained end-to-end for binary OUD diagnosis prediction.\cite{devlin2019bert}
The model takes as input a patient's longitudinal diagnosis history, represented as a temporally ordered sequence of clinical encounters. Within each encounter, all recorded ICD-10 diagnosis codes are included as discrete input tokens. 
A special query token [QENC] is prepended to the end of the sequence, and its final hidden state is passed through a linear classification head to produce a binary prediction of whether the patient will receive an OUD-related diagnosis at the next encounter. 
Segment embeddings are used to distinguish diagnoses belonging to different encounters, allowing the model to capture both the content and temporal ordering of a patient's clinical history.

Crucially, the input vocabulary is restricted to the top-$K$ diagnosis codes selected by each feature selection method under evaluation. 
The model is trained from scratch for each combination of feature selection method and vocabulary size $K$, ensuring that observed performance differences are attributable to the quality of the selected features rather than any pre-existing knowledge. 
This design provides a direct, outcome-driven benchmark for comparing feature selection strategies in the context of OUD prediction.

\subsection*{\textit{Evaluation Metrics}}
Model performance was assessed using three complementary metrics. The Area Under the Precision-Recall Curve (AUPRC) served as the primary metric, as OUD-positive cases constitute a minority class in the patient population and AUPRC is known to be more informative than AUROC under class imbalance. The Area Under the Receiver Operating Characteristic Curve (AUROC) was reported as a secondary metric to assess overall discriminative ability across all classification thresholds. We additionally report the Optimal F1 score, defined as the maximum F1 score achievable across all decision thresholds, which reflects the best attainable balance between precision and recall. All metrics were computed on a held-out test set and reported as the mean over 12 independent runs with different random seeds (42–53) to ensure robustness.

All evaluations were conducted on the held-out test set, and no test information was used during feature selection or model training.

\begin{table}[h]
\centering
\caption{Downstream prediction performance across feature selection methods and cross-method consensus sets at $K \in \{300, 500\}$. Consensus $\geq$3 and $\geq$4 denote feature subsets selected by at least 3 or 4 of the 5 methods, with actual feature counts shown.}
\label{tab:feature_selection_results}
\resizebox{\textwidth}{!}{%
\begin{tabular}{lcccccc}
\toprule
\textbf{Method} & \textbf{\# Features} & \textbf{BERT AUROC} & \textbf{BERT AUPRC} & \textbf{BERT Opt.\ F1} & \textbf{LogReg AUPRC} & \textbf{MLP AUPRC} \\
\midrule
\multirow{2}{*}{Recurrence Enrichment} & 300 & 0.867 & 0.191 & 0.221 & 0.045 & 0.132 \\
                                        & 500 & 0.846 & 0.191 & 0.225 & 0.045 & 0.132 \\
\multirow{2}{*}{NTK Sensitivity}        & 300 & 0.891 & 0.247 & 0.277 & 0.044 & 0.150 \\
                                        & 500 & 0.901 & 0.291 & 0.319 & 0.047 & 0.148 \\
\multirow{2}{*}{LightGBM-SHAP}          & 300 & 0.891 & 0.228 & 0.259 & 0.051 & 0.150 \\
                                        & 500 & 0.901 & 0.266 & 0.293 & 0.047 & 0.149 \\
\multirow{2}{*}{Elastic Net}            & 300 & 0.867 & 0.189 & 0.220 & 0.065 & 0.145 \\
                                        & 500 & 0.886 & 0.216 & 0.245 & 0.066 & 0.143 \\
\multirow{2}{*}{LLM Semantic}           & 300 & 0.832 & 0.182 & 0.216 & 0.043 & 0.135 \\
                                        & 500 & 0.844 & 0.191 & 0.224 & 0.045 & 0.138 \\
\midrule
\multirow{2}{*}{Consensus $\geq$3}      & 229 & 0.878 & 0.210 & 0.240 & 0.052 & 0.142 \\
                                        & 440 & 0.895 & 0.252 & 0.282 & 0.052 & 0.151 \\
\multirow{2}{*}{Consensus $\geq$4}      &  82 & 0.805 & 0.166 & 0.201 & 0.053 & 0.133 \\
                                        & 191 & 0.846 & 0.186 & 0.218 & 0.047 & 0.138 \\
\midrule
Full Vocabulary                         & 1{,}908 & 0.911 & 0.458 & 0.517 & 0.047 & 0.158 \\
\bottomrule
\end{tabular}}
\end{table}

\section*{Experiments and Results}

To systematically evaluate the impact of feature selection on downstream clinical prediction, all feature subsets were assessed under a controlled and unified experimental framework. The patient cohort was partitioned into train, validation, and test sets at the patient level, with splits fixed across all experiments to ensure fair comparability.

Five feature selection methods were evaluated: Recurrence Enrichment, NTK Sensitivity, LightGBM-SHAP, Elastic Net, and LLM Semantic Similarity. For each method, diagnosis codes were ranked as described in Section 2, and feature subsets were constructed using vocabulary budgets $K \in {50, 100, 150, 200, 300, 500}$. For each (method, $K$) combination, a BERT-based binary prediction model was trained from scratch using identical architecture, hyperparameters, and training protocol, with the input vocabulary restricted to the selected $K$ codes. A full-vocabulary BERT model trained on all available diagnosis codes served as an upper-bound reference.

To contextualize BERT performance, we additionally trained two conventional baseline classifiers—Logistic Regression and Multi-Layer Perceptron (MLP)—under the same feature subsets and vocabulary budgets, following standard practice in clinical prediction benchmarking.

Model performance was evaluated exclusively on the held-out test set. Bootstrap resampling was applied to the test cohort to compute 95\% confidence intervals for all reported metrics. Observed differences in predictive performance across feature selection methods therefore reflect the quality of the selected feature subsets, rather than variations in model architecture or data partitioning.

\begin{figure}[b]
\centering
\includegraphics[width=1.0\linewidth]{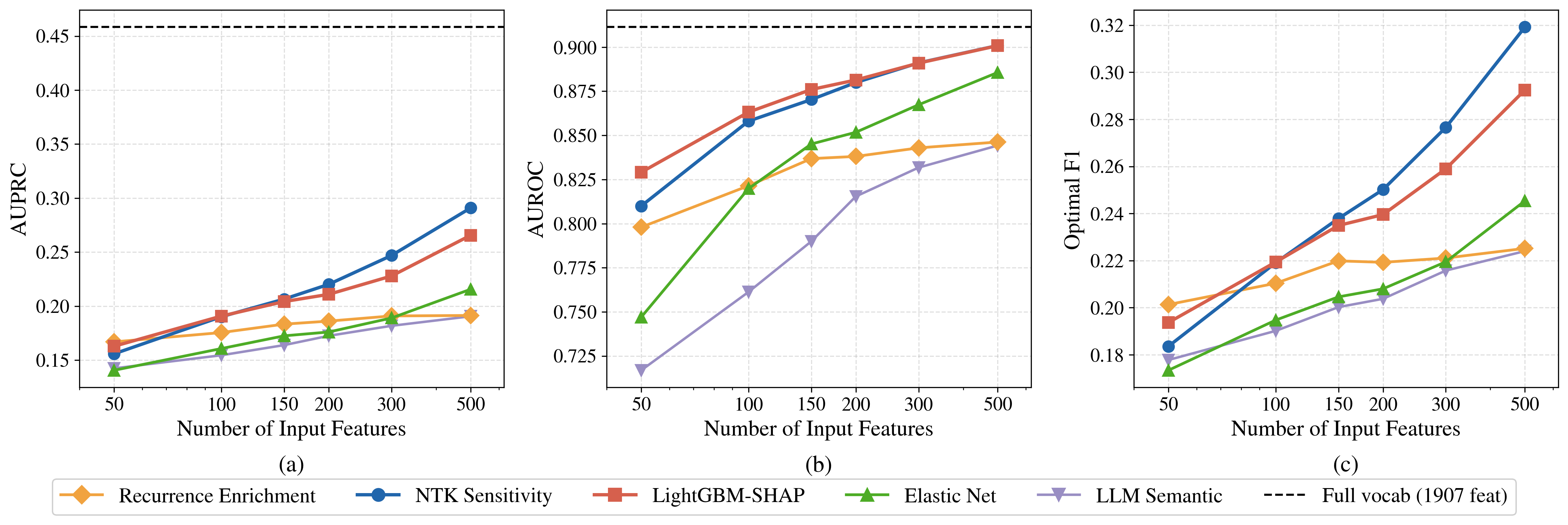}
\caption{Downstream prediction performance across different feature selection methods and feature set sizes.}
\label{fig:performance}
\end{figure}

\subsection*{\textit{Overall Predictive Performance}}
Table 1 summarizes predictive performance across feature selection methods at $K \in {300, 500}$. NTK Sensitivity achieved the highest overall BERT performance, reaching an AUPRC of 0.291, AUROC of 0.901, and Optimal F1 of 0.319 at $K = 500$. LightGBM-SHAP produced comparable performance (AUPRC = 0.266, AUROC = 0.901), while Elastic Net yielded intermediate results (AUPRC = 0.216). Recurrence Enrichment and LLM Semantic yielded the lowest AUPRC values among all methods (AUPRC $\approx$ 0.191 at $K = 500$). Across all feature budgets, the BERT classifier consistently outperformed both logistic regression and MLP baselines under equivalent feature subsets, underscoring the importance of model architecture in capturing temporal diagnostic patterns.

\subsection*{\textit{Top feature analysis}}

As shown in Figure\ref{fig:performance}, predictive performance improved consistently as the number of selected features increased from 50 to 300, with notable gains in AUPRC, AUROC, and Optimal F1 observed across all methods. Beyond approximately 300 features, performance gains became marginal, suggesting diminishing returns as the feature budget expanded toward 500. This plateau was consistent across all five feature selection methods, indicating that a moderate feature vocabulary of 300 codes may be sufficient to capture the most diagnostically informative signal for OUD prediction. The full-vocabulary upper bound (1,908 features; AUPRC = 0.458) nonetheless remained substantially higher than all feature-selected configurations, suggesting that further performance gains may be achievable with larger feature budgets.

\subsection*{\textit{Stability Analysis}}

\begin{wrapfigure}{r}{0.58\textwidth}
  \centering
    \includegraphics[width=0.58\textwidth]{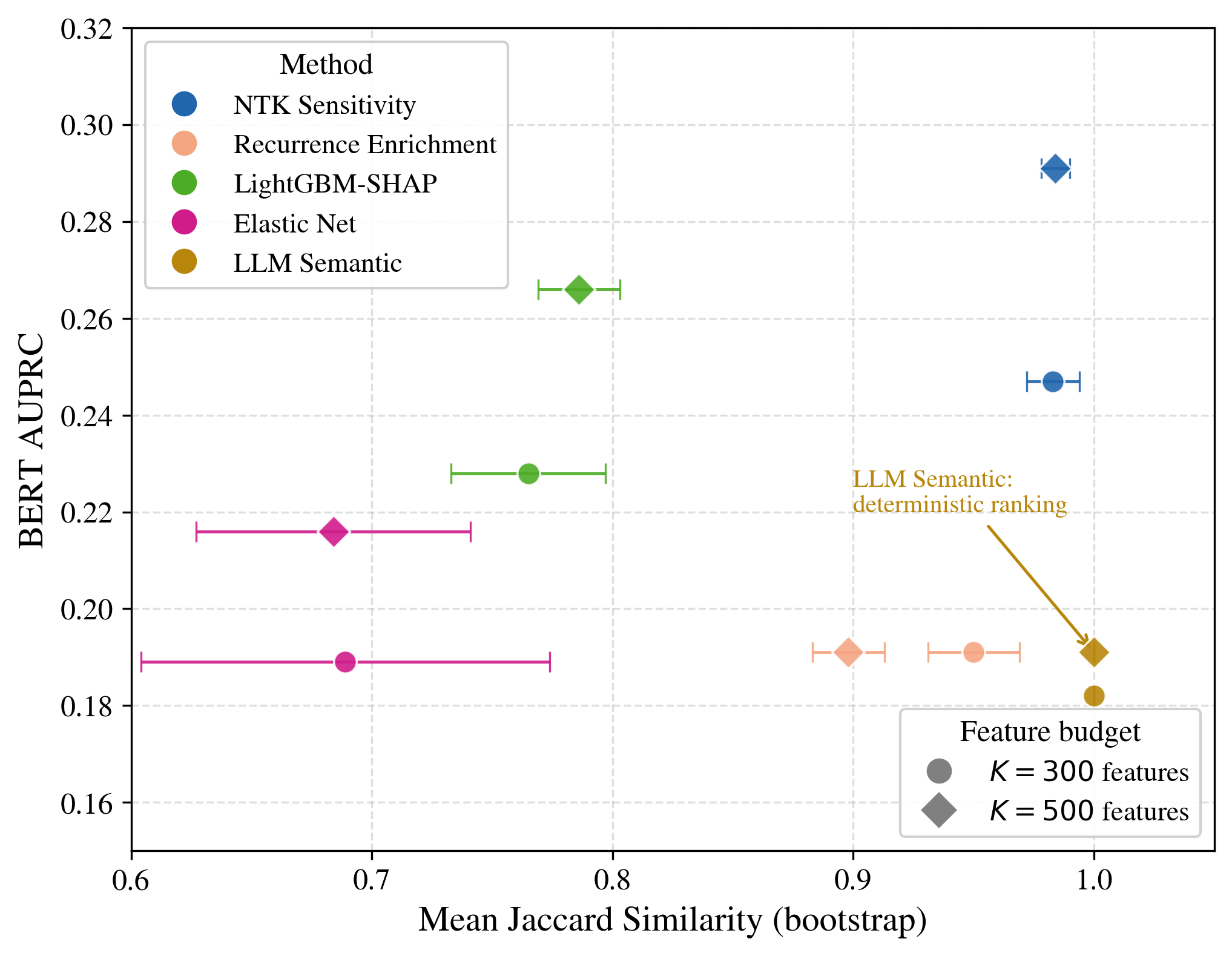}
 \caption{Stability and agreement of selected features across feature selection methods.}
 \label{fig:bootstrap}
\end{wrapfigure}

We examined the stability of selected feature subsets using bootstrap resampling of the training data, measuring pairwise Jaccard similarity of the top-$K$ feature lists across resamples (Figure \ref{fig:bootstrap}). NTK Sensitivity exhibited near-deterministic behavior, achieving mean Jaccard similarity of $0.983 \pm 0.011$ at $K = 100$ and $0.984 \pm 0.006$ at $K = 300$, indicating that its rankings are highly robust to perturbations in training data composition. Recurrence Enrichment also demonstrated high stability ($0.950 \pm 0.019$ at $K = 100$), though stability decreased modestly at larger vocabulary sizes ($0.898 \pm 0.015$ at $K = 300$). LightGBM-SHAP showed moderate stability ($0.765 \pm 0.032$ at $K = 100$; $0.786 \pm 0.017$ at $K = 300$). Elastic Net demonstrated substantially lower and more variable stability ($0.689 \pm 0.085$ at $K = 100$; $0.684 \pm 0.057$ at $K = 300$), likely reflecting sensitivity to correlated diagnosis features under $\ell_1/\ell_2$ regularization.




\subsection*{\textit{Cross-Method Overlap}}


We also calculated the overlap among the top-100 features selected by each method. Substantial intersections were observed among NTK Sensitivity, LightGBM-SHAP, and Recurrence Enrichment, with 60, 56, and 38 shared codes respectively across pairwise comparisons, suggesting convergent identification of core OUD-associated diagnostic patterns. The cross-method consensus: 61 diagnosis codes were selected by at least three methods, and 23 codes were shared by at least four methods, forming a robust consensus feature set. LLM Semantic introduced a complementary set of features not consistently identified by purely data-driven methods, with relatively low pairwise overlap with all other approaches (Jaccard $\leq 0.156$), suggesting that semantic reasoning captures clinically relevant diagnoses that frequency and gradient-based methods may overlook.


\section*{Discussion}

In this work, we performed a controlled benchmark of feature selection for high-dimensional, sparse diagnosis-code representations in EHR-based OUD prediction. By implementing five feature selection methods under a unified preprocessing and evaluation protocol, our study contributes (i) an end-to-end comparison across statistical, machine-learning, and LLM-assisted selection strategies, (ii) an outcome-driven assessment using a fixed downstream modeling setup, and (iii) an analysis that goes beyond discrimination to examine resampling stability and cross-method agreement. Empirically, we observed consistent performance gains as the feature budget increased from $K=50$ to roughly $K=300$, followed by diminishing returns, suggesting that a moderate diagnosis vocabulary can capture much of the predictive signal under realistic deployment constraints. Among the methods, NTK-motivated early gradient sensitivity achieved the strongest overall performance while also exhibiting near deterministic stability, indicating that early sensitivity signals can yield robust rankings in sparse, correlated diagnosis spaces. Tree-based importance (LightGBM--SHAP) achieved competitive performance but showed only moderate stability, while Elastic Net exhibited larger variability, consistent with feature competition under collinearity. Finally, LLM-guided selection underperformed as a standalone approach in this cohort, but its low overlap with data-driven rankings suggests a potentially complementary role as an external semantic prior for candidate generation or re-ranking rather than replacement.

Several limitations should be considered. First, the outcome definition based on diagnosis codes may be affected by under-coding, documentation variability, and temporal misalignment with true onset, motivating sensitivity analyses using alternative labeling strategies and incident-only definitions. Second, ICD-9$\rightarrow$ICD-10 harmonization and three-character truncation improve comparability and reduce sparsity, but can collapse clinically distinct subphenotypes; future work could test hierarchy-aware selection that preserves granularity only where it adds predictive value. 
Third, we focused on diagnosis codes alone. In future work, we plan to incorporate additional EHR domains, including medications, procedures, laboratory results, and social determinants of health, as these may affect both the optimal feature budget and the relative advantages of different feature-selection paradigms.
Finally, although this benchmark was conducted on the HealthFacts database, external validation across health systems and time periods is needed to assess the transportability of selected feature sets and their downstream impact.

\section*{Conclusion}

We presented a unified empirical comparison of five feature selection paradigms for high-dimensional, sparse diagnosis-code features in EHR-based OUD prediction. Across a range of feature budgets, we found that predictive performance improves rapidly up to a moderate vocabulary size (approximately 300 diagnosis categories) and then exhibits diminishing returns. Among the evaluated strategies, NTK-motivated early gradient sensitivity achieved the best overall downstream performance while also demonstrating exceptionally high stability under resampling, making it a strong candidate for reproducible feature selection in sparse clinical settings. Tree-based SHAP attributions provided competitive accuracy but with lower stability, whereas Elastic Net and univariate enrichment exhibited larger trade-offs between performance and reproducibility. LLM-guided semantic selection produced feature sets with limited overlap with data-driven methods, suggesting a complementary role as an external semantic prior rather than a standalone selector. Collectively, our findings provide practical guidance for selecting compact, informative, and stable diagnosis feature sets for EHR modeling under realistic computational and deployment constraints.


\section*{Acknowledgment}
This work was supported by the Patient-Centered Outcomes Research Institute (PCORI) under Contract No. ME-2023C3-35532.
\makeatletter
\renewcommand{\@biblabel}[1]{\hfill #1.}
\makeatother

\bibliographystyle{vancouver}
\bibliography{amia}  

\end{document}